# On some limitations of current data-driven weather forecasting models

Massimo Bonavita

*ECMWF, Reading, UK*

## ABSTRACT

As in many other areas of engineering and applied science, Machine Learning (ML) is having a profound impact in the domain of Weather and Climate Prediction. A very recent development in this area has been the emergence of fully data-driven ML prediction models which routinely claim superior performance to that of traditional physics-based models. In this work, we examine some aspects of the forecasts produced by an exemplar of the current generation of ML models, Pangu-Weather, with a focus on the fidelity and physical consistency of those forecasts and how these characteristics relate to perceived forecast performance. The main conclusion is that Pangu-Weather forecasts, and possibly those of similar ML models, do not have the fidelity and physical consistency of physics-based models and their advantage in accuracy on traditional deterministic metrics of forecast skill can be at least partly attributed to these peculiarities. Balancing forecast skill and physical consistency of ML-driven predictions will be an important consideration for future ML models. However, and similarly to other modern post-processing technologies, the current ML models appear to be already able to add value to standard NWP output for specific forecast applications and combined with their extremely low computational cost during deployment, are set to provide an additional, useful source of forecast information.

# 1. Introduction

As it has been the case in disparate areas of applied science and engineering, Machine Learning (ML) methods are having a profound and pervasive impact in Numerical Weather Prediction (NWP) and Climate monitoring and prediction (e.g., Bonavita et al., 2023; Schneider et al., 2022; Bonavita et al., 2021, for recent overviews). The vast majority of applications of ML methods in NWP and Climate have sought to deploy ML algorithms in specific parts of the NWP and Climate prediction chain, aiming to take advantage of the extremely low computational cost of the deployed ML models and the fact that the ML algorithms can be effective at learning complex, nonlinear mappings if large and accurate datasets are available for their training. Examples include the use of ML algorithms for observation pre-processing and quality control, data assimilation, emulation of observation operators, emulation of model components, development/improvement of new model parameterisations, post-processing of NWP and Climate prediction outputs (see e.g., Krasnopolsky, 2023, for a recent review of some of the main research areas). The route of adapting ML tools to specific aspect of the NWP workflow has been the one typically favoured by practitioners of the field as it facilitates understanding and interpretability of the ML model outputs, which is crucial for user uptake and also for long-term continued development (e.g., McGovern et al., 2023).

In the last few years, a parallel and growing development area has emerged which aims to use ML methods to produce fully data-driven forecast models for NWP and Climate prediction. These efforts have been made possible by the availability of high-quality, multi-decadal Earth system reanalysis products, such as the ECMWF ERA5 reanalysis (Hersbach et al., 2020), which have provided the ML community with readily accessible, curated, and accurate datasets necessary for training of the ML models. The first notable results in this area have been achieved by Keisler, 2022, where the trained ML model shows deterministic forecast skill scores which are competitive with NOAA Global Forecast System (GFS) operational forecasts and comparable with ECMWF Integrated Forecast System (IFS) operational forecasts. Compared to earlier efforts, Keisler's model was trained on a dataset of ERA5 reanalysis fields at significantly higher horizontal (1 degree lat/lon regular grid) and vertical (13 pressure levels, from 50 to 1000hPa) resolution every 6 hours, with the explicit aim of learning the set of physical laws driving the ECMWF IFS and other traditional NWP models. After the appearance of Keisler's work, various independent groups, often affiliated with large technology corporations, have announced the development of fully data-driven ML weather forecast models and published initial evaluations of their performance. Notable examples include FourCastNet (Pathak et al., 2022); Pangu-Weather (Bi et al., 2022, 2023); SwinRDM (Chen et al., 2023a); ClimaX (Nguyen et al., 2023); GraphCast (Lam et al., 2022), FengWu (Chen et al., 2023b). While these ML models show variations in their architecture and training, some common fundamental themes are apparent. With respect to the original Keisler's model, all these models are built using vastly larger training datasets obtained by sampling the ERA5 reanalysis fields at higher horizontal/vertical/temporal resolution (typical values are 0.25 lat/lon regular grid, 13 to 37 vertical pressure levels, 1 to 6-hour temporal sampling). This is a deliberate design choice which is justified by the goal of increasing the realism and fidelity of the weather features the ML model is able to predict, though at the cost of vastly increasing the memory footprint of the model, which can become a limiting factor during training (Lam et al., 2022). At the same time, exploiting the increased dimensionality of the data has the effect of requiring a ML model with a vastly larger number of parameters ($O(10\text{-}100)$) than in the original Keisler's model, which used a Graph Neural Network architecture with 6.7 million trainable parameters. The large number of trainable parameters can become an issue in terms of computational cost and performance during the training phase of the ML model (Bi et al., 2022).

Together with increased size of the training dataset, ML model complexity, memory footprint and computational training costs, the forecast performance of the ML models has improved. All the most recent ML models claim to be able to outperform the ECMWF IFS system (currently the most accurate physics-based global NWP forecasting system in the world) on a variety of performance metrics for deterministic, single-shot prediction (Bi et al., 2022, 2023; Chen et al., 2023a, b; Lam et al., 2022). Together with the strikingly low computational cost and energy consumption of the ML model during the deployment phase with respect to standard NWP models ($O(10^4)$ faster and computationally/energy cheaper), claims that the era of traditional NWP is rapidly coming to an end in favour of a new generation of ML-driven Weather Prediction are becoming common (MLWP; Bi et al., 2022, 2023; Chen et al., 2023; Lam et al., 2022). Notwithstanding the tumultuous progress of MLWP in the past few years, we consider here the question of to what extent and with which caveats, if any, these claims are justified, building on the initial evaluation in Ben-Bouallegue et al., 2023, but with a specific focus on the spatial and physical realism of the MLWP forecast fields. The approach we have taken here is to choose a representative sample of the current generation of MLWP model (Pangu-Weather, Bi et al., 2022, 2023) and analyse some aspects of its output from the point of view of physical consistency and fidelity and how these characteristics affect perceived forecast performance. We have done this exercise by comparing the characteristics of Pangu-Weather forecasts to those of the ERA5 reanalysis fields used in the training and the operational ECMWF IFS physics-based model which is the NWP modelling system against which the forecast performance of Pangu-Weather and other MLWP models is typically validated. This comparison is obviously not exhaustive but aims to highlight differences between a standard NWP model and a representative MLWP model that we believe are significant from a meteorological and forecasting perspective.

Outline of the paper is as follows. In section 2 we provide a brief description of the Pangu-Weather model. In section 3 we provide diagnostics of the behaviour of the Pangu-Weather model from an energy spectral decomposition perspective and discuss their consequences on the Pangu-Weather forecast characteristics. In section 4 we look at the balance between mass and wind fields as represented in the Pangu-Weather model, ERA5 and ECMWF IFS. In section 5 we look at forecast performance of Pangu-Weather and ECMWF IFS for selected variables. In sections 6 we discuss the main findings of this work, their relevance in the current debate on NWP and MLWP and a personal outlook on the possible evolution of MLWP.

## 2. A representative MLWP model: Pangu-Weather

Pangu-Weather (Bi et al., 2022, 2023) is a Deep Learning (DL) forecast model trained on 43 years

(1979-2021) of ERA5 reanalysis data developed by researchers at Huawei Cloud Computing. ERA5 global analyses are retrieved hourly at 0.25 deg regular lat/lon horizontal resolution and 13 pressure levels, plus a small selection of surface fields (T2m, u/v10m, mslp). The architecture of Pangu-Weather is based on a variation of the Transformer model (Vaswani et al., 2017) which is widely adopted in large language models (GPT-3, BERT), and, more specifically, its adaptation to Computer Vision tasks (Vision Transformers, Dosovitskyi et al., 2021). The Pangu-Weather model is one of the largest in terms of number of parameters (~256M), possibly due to the quadratic complexity in the number of tokens induced by the Transformer architecture, which has the effect of requiring extended and computationally intensive training (Bi et al., 2022, sec. 3.4) and the need to train separate ML models for different forecast ranges (see below).

An interesting peculiarity of Pangu-Weather is the technique used for producing forecasts at different lead times. Most other ML models predict the evolution of the atmosphere with a Δt=6 hours timestep future and forecasts at longer lead times, which are always multiple integers of Δt, are obtained autoregressively:

$$X^{t+T} = ML(X^{t+T-\Delta t}) \quad (1)$$

As noted by the Pangu-Weather developers and others, the repeated application of an imperfect model leads to rapid accumulation of errors and can drastically limit the predictive skill of the model. Other ML models obviate the problem by progressively increasing the forecast time over which the ML model is optimised to minimise forecast errors (e.g., Lam et al., 2022), typically at the price of increased blurriness of the forecasted states. The solution adopted in Pangu-Weather is called "Hierarchical Temporal Aggregation" (HTA) and effectively involves the development of 4 separate ML models trained to forecast at different lead times of 1, 3, 6 and 24 hours. The advantages of the technique are the reduction of the number of applications of the ML model for any given forecast lead time, which may reduce forecast error growth; and the ability to provide forecasts with 1-hour granularity, which is the highest possible given that ERA5 fields are archived hourly. A possible disadvantage is that applying different ML models in a forecast can lead to unphysical discontinuities in forecast evolution.

Another aspect where Pangu-Weather differs from the other published ML forecast models is that it uses a mean absolute error loss function (L1 norm) in the training instead of the more common mean squared error (L2) norm. The choice is justified as beneficial for the convergence speed of the training, which may be a consequence of the general property of L1 loss functions of encouraging sparsity in regression problems (e.g., Hsieh, 2023).

The Pangu-Weather model is publicly available (https://github.com/198808xc/Pangu-Weather) for non-commercial use and has been used to build a six month dataset of forecasts (October 2018 to March 2019, every three days; Matthew Chantry, ECMWF, personal communication) started from both ERA5 analysis fields and ECMWF IFS analyses. This dataset has been used in this study, together with publicly available ERA5 reanalysis fields and ECMWF IFS analyses and forecasts over the same period. Additionally, a selection of MLWP models have been run in a semi-operational configuration at ECMWF since August 2023 (graphical outputs available at https://charts.ecmwf.int). Some of these daily graphical products have also been used in this work.

## 3. Spectral diagnostics of Pangu-Weather, ERA5 and ECMWF IFS

A widely identified issue with forecasts produced by ML models is that they appear to become increasingly "blurry" with increasing forecast lead times (Keisler, 2020; Lam et al., 2022). This behaviour of the ML models is to be expected on general grounds (Sønderby et al. 2020), as they are usually trained to optimize a weighted mean squared/absolute error (L2/L1 norm). Thus, the way the ML models express increasing uncertainty over longer lead times is by producing forecasts closer to the forecast mean of the forecast error pdf, which will result in progressively smoother forecast states. This is indeed what happens in the Pangu-Weather model as well. In Figure 1 we show four examples of energy power spectra from the ERA5 analysis, Pangu-Weather forecasts and ECMWF IFS forecasts at different lead times. For consistency, ERA5 analyses and ECMWF IFS forecasts were first interpolated to the 0.25 deg regular lat/lon grid used by Pangu-Weather and then transformed to spectral space. First thing to notice is that the ECMWF IFS forecast spectra do not change appreciably with forecast lead time and remain close to the ERA5 analysis spectra until about wave number 200 (~200 km wavelength), above which they start to diverge (being more energetic) likely due to the higher spatial resolution of the ECMWF IFS forecasts (~ 9km horizontal grid spacing vs ~31 km of ERA5 analyses) and, to a lesser extent, the impact of employing models from different IFS cycles (IFS Cycle 45r1 for the ECMWF IFS versus IFS Cycle 41r2 for ERA5). Conversely, the spectra of Pangu-Weather forecasts show a noticeable divergence from the ERA5 analysis spectra in terms of reduced energy already from wavenumbers in the 60-80 range (~500-700 km wavelength). Additionally, this reduction in the spectral energy of the Pangu-Weather forecasts shows a marked sensitivity to forecast lead time, especially noticeable for the t+24h forecasts. For completeness, note that deterministic forecasts from ERA5 re-analyses (hindcasts) at the original resolution and IFS model cycle are also freely available. Their forecast spectra are not shown here because they are largely indistinguishable from the spectra of ERA5 analysis fields.

This confirms that Pangu-Weather, like other ML models, produces less spectrally resolved forecasts than the analysis fields used in their training and even less than those produced by the ECMWF IFS forecasts which are typically used to verify against. In other words, the effective resolution of Pangu-Weather forecasts is closer to 500-700 km than to the nominal 0.25 deg and is decreasing with forecast lead time, most notably during the first 24 hours. The Hierarchical Temporal Aggregation idea of training and deploying different models for different lead times seems however effective in reducing the further loss of spectral energy beyond t+24h lead time.

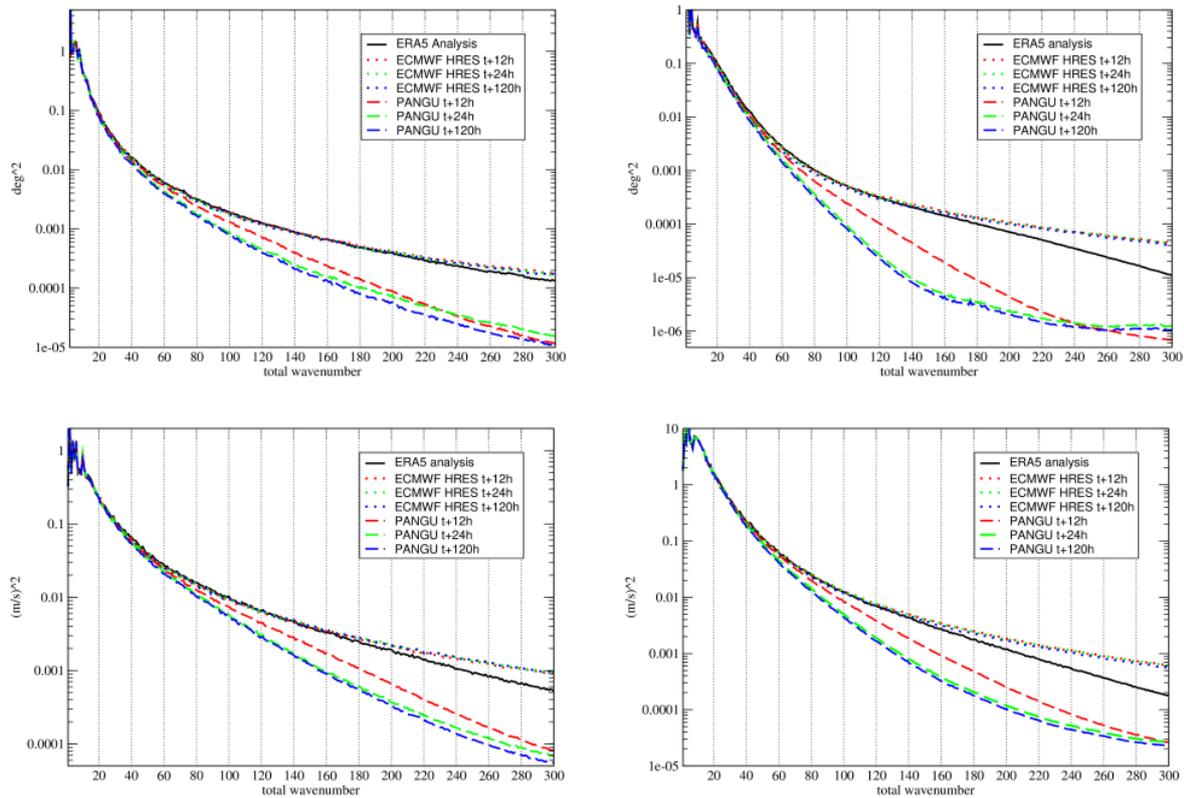

Figure 1: *Power spectral density as a function of total wavenumber of ERA5 analysis (continuous line), ECMWF IFS operational forecasts (dotted lines) and Pangu-Weather forecasts (dashed lines) at lead times t+12h, t+24h and t+120h. Top row: temperature field at 850hPa (left panel) and 250 hPa (right panel); Bottom row: wind speed field at 850hPa (left panel) and 250 hPa (right panel).*

What are the consequences of the rapid reduction in the Pangu-Weather forecast spectral energy with increasing wavenumber? Looking at standard forecast maps used for synoptic evaluation, the differences between Pangu-Weather and ECMWF IFS forecasts are not striking, e.g., Figure 2, although the Pangu-Weather contours appear somewhat smoother at closer inspection.

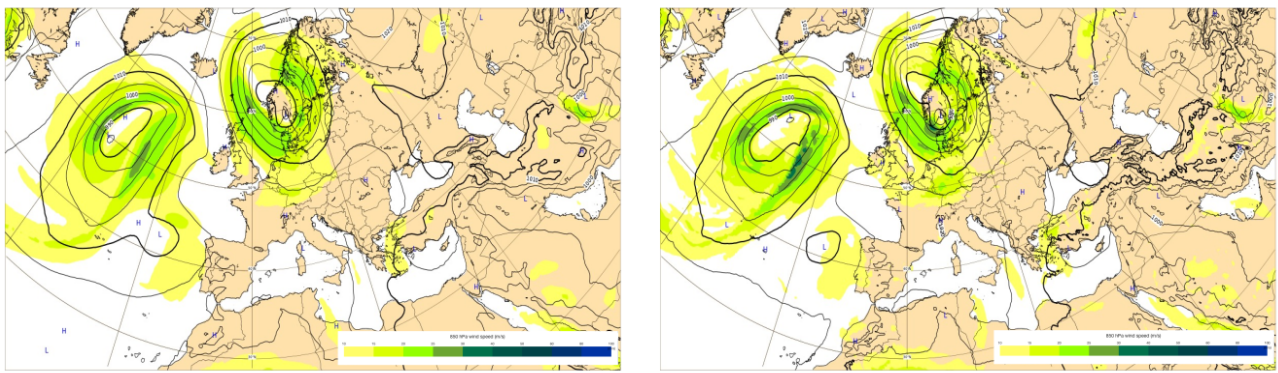

Figure 2: *T+72 hour forecasts of mean sea level pressure (continuous lines, 5 hPa intervals) and wind speed at 850 hPa from Pangu-Weather (left panel) and the ECMWF IFS (right panel) valid on 06-08-2023 at 00 UTC.*

However, in forecasting weather phenomena which have high variability on sub-synoptic and mesoscale spatial modes, the characteristics of the Pangu-Weather model can become more noticeable. An example is given in Figure 3, where we present five-day (t+132h) forecasts of the evolution of tropical cyclone Doksuri. Typhoon Doksuri was a category 4 tropical cyclone which caused extensive damage in the Philippines, Taiwan, China and Vietnam, in late July 2023. It is visually apparent how typhoon Doksuri evolves in the Pangu-Weather forecast to become a shallow low pressure system (986 hPa min mslp), while it remains an active tropical cyclone in the ECMWF IFS forecast (957 hPa min mslp), though still not as deep as it was in reality (944 hPa, IBTrACS version v04r00, Knapp et al., 2018). Based on the spectral characteristics of the Pangu-Weather model forecasts, this behaviour is to be expected and leads to performance in the forecast of the intensity of tropical cyclones which is not as good as that of state-of-the-art NWP Earth system simulators like the IFS (see Ben-Bouallegue et al., 2023 for a detailed discussion).

**3.1 Is Pangu-Weather emulating the ECMWF ensemble forecast mean?**

As mentioned earlier, a common interpretation of the progressive loss of detail for increasing forecast ranges of ML data-driven models is that these models are effectively trained to optimise a mean squared/absolute error (L2/L1) norm of forecast errors at increasing lead times. This is equivalent, for normally distributed variables, to estimate the posterior mean/median of the forecast distribution conditioned on the set of input fields used in the training. As the forecast pdf becomes broader with increased forecast lead time, the ML forecast will express the growing uncertainty by progressively smoothing out unpredictable details from the forecast. It is thus of interest to see if the Pangu-Weather forecasts share some of the features of the forecast mean fields produced by the ECMWF ensemble prediction system (ENS; ECMWF, 2022). The ECMWF ENS system is expressly set up to sample the forecast pdf of the ECMWF IFS starting from a sample of the initial uncertainties estimated by the ECMWF Ensemble of Data Assimilation (EDA; Bonavita et al., 2016)

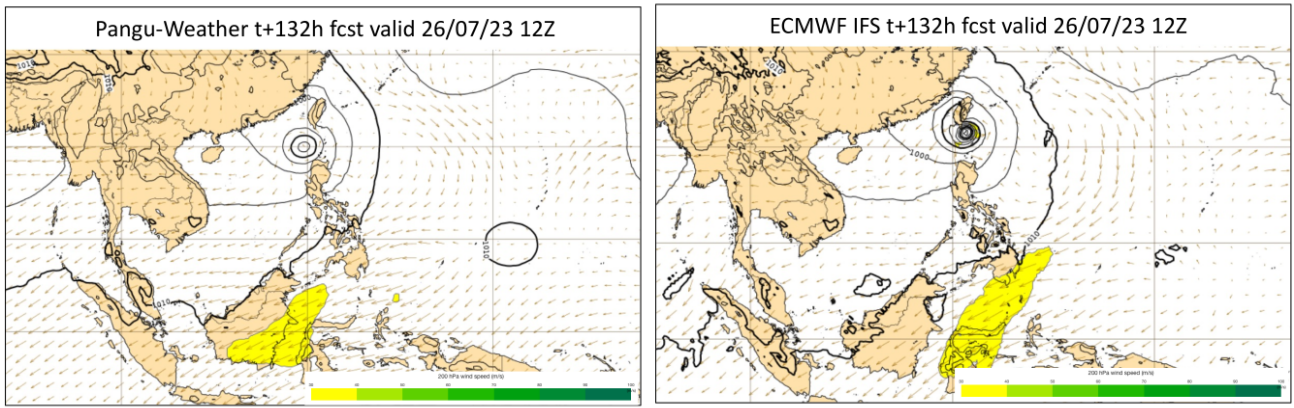

Figure 3: *T+132 hour forecasts of mean sea level pressure (continuous lines, 5 hPa intervals) and wind speed at 200 hPa for Pangu-Weather (left panel) and the ECMWF IFS (right panel) valid on 26-07-2023 at 012 UTC in the area covering typhoon Doksuri.*

In Figure 4 we present a similar set of plots as in Figure 1 but showing curves for the energy spectra of the ECMWF ENS ensemble forecast mean (EM) instead of the deterministic ECMWF IFS forecast. From these plots it is immediately apparent that the spectral signature of the ECMWF EM forecasts is quite different from that of the Pangu-Weather forecasts. In the short forecast range (12-24 hours), where error evolution on synoptic and sub-synoptic scales is approximately linear, the EM spectra closely track the ERA5 analysis spectra, while the Pangu-Weather forecasts already show the signature of heavy damping of spectral modes above approx. wavenumber 60. On the other hand, at forecast lead time well into the medium range (t+120h), the ECMWF EM fields show reduced energy at synoptic ranges, where error growth becomes nonlinear and ensemble averaging acts to smooth out unpredictable forecast modes (Leith, 1974; Toth and Kalnay, 1997), while they maintain a comparable or larger spectral signature than Pangu-Weather forecasts for higher spatial frequency modes above approx. 120-140.

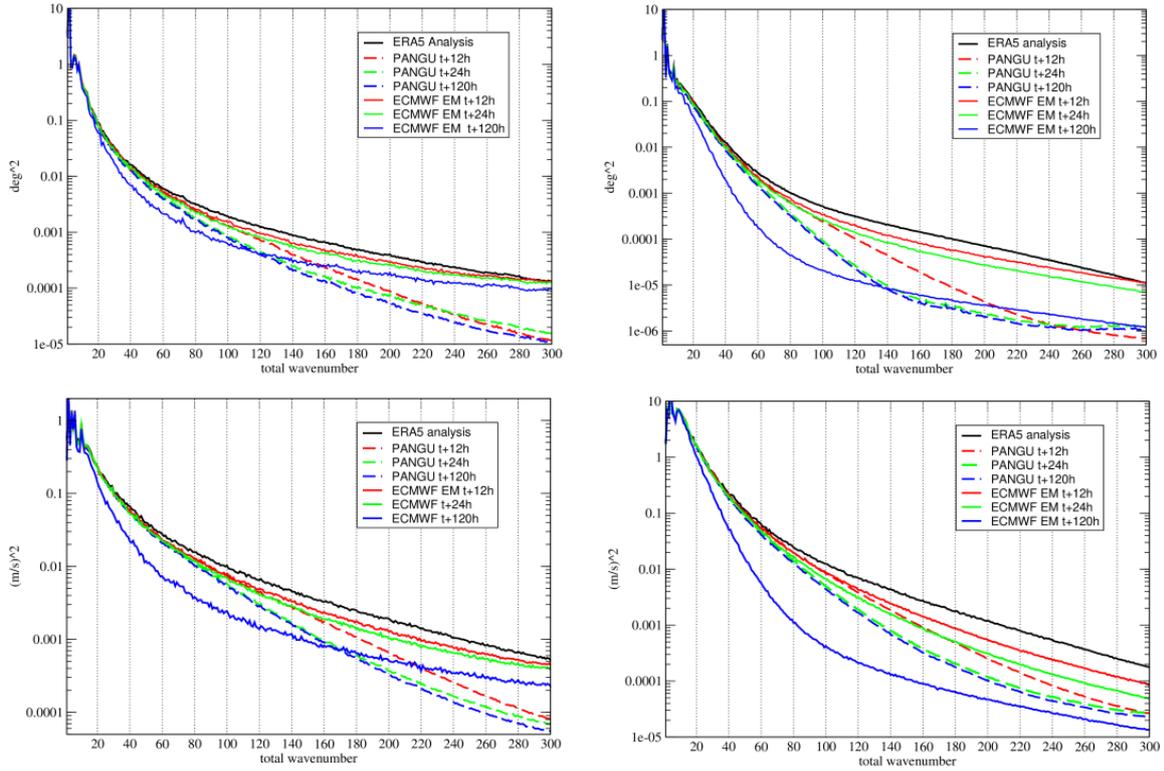

Figure 4: *Power spectral density as a function of total wavenumber of ERA5 analysis (continuous black line), ECMWF IFS operational ensemble forecast mean (EM, continuous lines of different colours) and Pangu-Weather forecasts (dashed lines) at lead times t+12h, t+24h and t+120h. Top row: temperature field at 850hPa (left panel) and 250 hPa (right panel); Bottom row: wind speed field at 850hPa (left panel) and 250 hPa (right panel).*

These results indicate that Pangu-Weather forecasts differ from ECMWF EM forecasts in the sense that they have both too little energy at short forecast ranges and high wave numbers, and too much energy in the medium range at synoptic and sub-synoptic scales. This is important in terms of the implications of employing the Pangu-Weather model in ensemble forecasting configuration and also for the interpretation of the results on Pangu-Weather forecast skill (Sec. 5).

## 4. Physical balances in Pangu-Weather

### 4.1. Geostrophic wind balance

Standard NWP models predict the evolution of the atmosphere by solving discretised forms of the governing physics-based equations (Pu and Kalnay, 2018). This set of equations describe fundamental conservation laws (momentum, mass, energy and constituents) that the atmospheric system obeys, and which implicitly enforce balances between the different variables that describe the atmosphere. One of the fundamental physical balances is the one between the mass variables (temperature, geopotential) and the wind field. Neglecting friction and acceleration, scale analysis (Holton, 2004) leads to a stationary diagnostic geostrophic balance between horizontal wind and geopotential in local Cartesian coordinates:

$$\mathbf{V}_g = \frac{1}{f}\hat{\mathbf{k}} \times \nabla_p \Phi \quad (2)$$

Where $\mathbf{V}_g = (u_g, v_g)$, $f = 2\Omega sin(\varphi)$ is the Coriolis parameter, and $\nabla_p \Phi$ is the gradient of the geopotential on an isobaric surface. The geostrophic balance is a good approximation for extra-tropical synoptic and larger scale flows, and it is interesting to see to what degree Pangu-Weather forecasts respect this balance.

In Figure 5 we present vertical profiles of the intensity of geostrophic wind, ageostrophic wind ($\mathbf{V}_{ag} \equiv \mathbf{V} - \mathbf{V}_g$) and the ratio of the intensity of ageostrophic over geostrophic winds. It is remarkable how the ECMWF IFS model remains close to the ERA5 analysis for all three quantities and at all lead times (in fact the ageostrophic wind shows a modest increase around the tropopause and lower stratosphere at longer lead times, possibly due to spin up effects). On the other hand, the Pangu-Weather forecasts profiles show reduced intensity of both geostrophic and ageostrophic winds, which also tend to further reduce for increasing lead time.

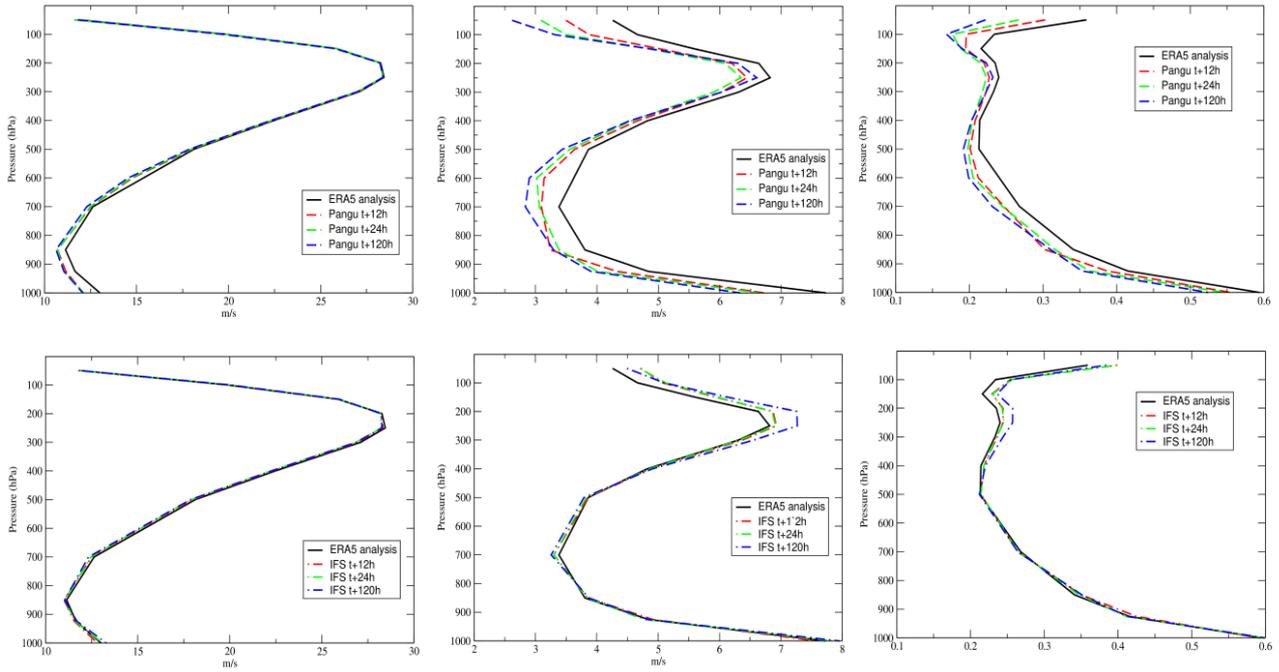

Figure 5: *Vertical profiles of the intensity of geostrophic wind (first column), ageostrophic wind (second column) and ratio of the intensity of ageostrophic over geostrophic wind (third column) over the extra-tropics ($|lat| \geq 20\ deg$) for the ERA5 reanalysis (continuous black line), Pangu-Weather forecasts (first row, dashed lines) and ECMWF IFS forecasts (second row, dot-dashed lines) at lead times t+12h, t+24h and t+120h. Values averaged over the 2023-09-07 to 2023-09-09 period.*

Additionally, this reduction in geostrophic and ageostrophic wind intensity is not balanced, i.e., the ratio of ageostrophic over geostrophic wind intensity is smaller than that diagnosed from the ERA5 analysis (and ECMWF IFS forecasts) and further decreasing with lead time. This implies that the wind and geopotential forecasts from the Pangu-Weather model are increasingly dynamically inconsistent with one another with forecast lead times.

Ageostrophic winds in midlatitude synoptic systems are connected with areas of convergence/divergence which, through the continuity equation, are linked to areas of vertical motions and active weather/precipitation. As shown in the example in Figure 6, while the broad scale geopotential pattern of the Pangu-Weather forecast appears plausible, the intensity of the ageostrophic motions (and by implication, the diagnosed vertical motions and precipitation) is not.

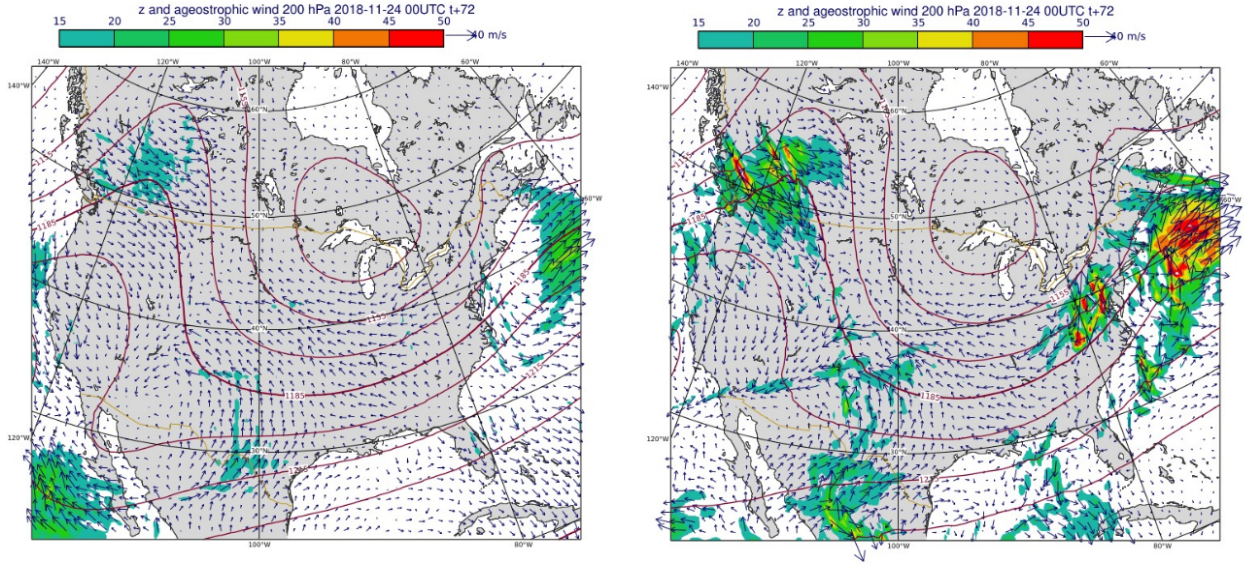

Figure 6: *T+72 forecasts of geopotential height (continuous lines, units dam), and ageostrophic wind (wind arrows and shaded areas for intensity, units m/s) at 200 hPa valid on 2018-11-24 at 00 UTC. Left panel: Pangu-Weather forecast. Right panel: ECMWF IFS forecast.*

**4.2. Rotational and divergent wind components**

The Helmholtz decomposition (Dutton, 1976) allows to uniquely partition the total circulation $\boldsymbol{u} = (u, v)$ into divergent ('vorticity free') and rotational ('divergence free') components:

$$\boldsymbol{u} = \boldsymbol{u}_d + \boldsymbol{u}_d = -\nabla\chi + \mathbf{k} \times \nabla\psi \quad (3)$$

where χ is the velocity potential function, and ψ is a streamfunction. χ and ψ can be obtained from the divergence (δ) and vorticity (ζ) fields, i.e.:

$$\nabla^2\chi = \delta, \quad \nabla^2\psi = \zeta. \quad (4)$$

Divergence and vorticity fields thus allow to estimate the dynamical consistency of the forecasted wind, in a manner analogous to what the geostrophic balance allows to do in terms of the dynamical consistency of mass and wind fields.

In Figure 7 we present vertical profiles of the ratio of the globally averaged absolute values of the divergence and vorticity fields for the ERA5 analysis, Pangu Weather forecasts, ERA5 forecasts and IFS operational forecasts. It is striking how this ratio is significantly and progressively reduced with increased forecast lead time in Pangu Weather forecasts, while remains approximately constant and close to that of the ERA5 analysis in both ERA5 forecasts (which are IFS forecasts run at the same resolution and with the same IFS model cycle used for the ERA5 reanalysis production) and operational IFS forecasts. The divergent component of the flow is significantly suppressed in Pangu Weather forecasts with respect to the rotational component (and also in absolute terms, not shown here), which is unphysical and also implies that diagnosed vertical motions would be suppressed, as discussed in the next section.

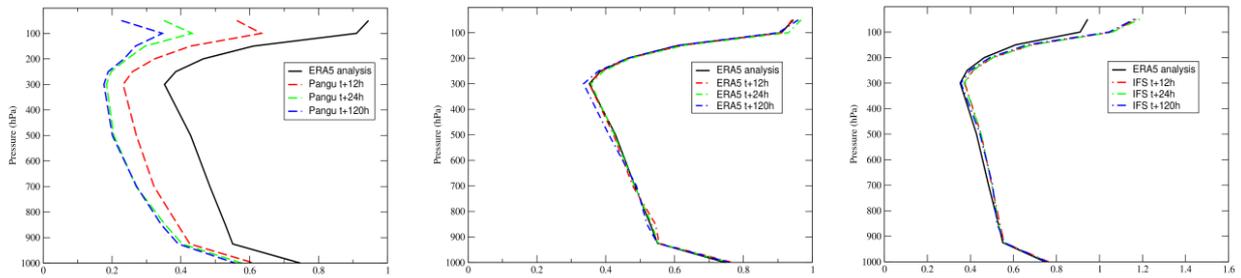

Figure 7: *Vertical profiles of the ratio of the globally averaged absolute values of divergence over vorticity for: ERA5 analysis (continuous line) and Pangu Weather forecasts (dashed line), left panel; ERA5 analysis (continuous line) and ERA5 forecasts (double dashed lines), middle panel; ERA5 analysis (continuous line) and ECMWF IFS forecasts (dot-dashed lines), right panel.*

**4.3. Vertical motions**

Like other MLWP models Pangu-Weather does not explicitly produce a forecast of vertical velocities. However, it is possible, under certain assumptions, to diagnose vertical velocity fields from horizontal velocity fields on constant pressure levels by integrating the continuity equation in the vertical (Holton and Hakim, 2013, Sec. 3.5.1):

$$\omega(p) = \omega(p_s) - \int_{p_s}^{p} div(\boldsymbol{u})_p dp \quad (5)$$

$$\omega(p) \cong -\rho g w(p) \quad (6)$$

As the geostrophic wind is approx. non divergent, except for the small effect due to the variation of the Coriolis parameter, vertical velocity can be diagnosed from the mean layer horizontal divergence, together with the standard hydrostatic balance assumption. We have applied this diagnostic using forecasted horizontal velocity fields from Pangu-Weather, ERA5 hindcasts and IFS on standard pressure levels in the 1000-500 hPa layer. An example is given in Figure 8, where we show the vertical velocity field ($w$) at 500 hPa forecasted by the ERA5 hindcast on a specific date (top panel); the vertical velocity field at 500 hPa diagnosed using Eqs. 5,6 from the ERA5 hindcast (middle panel); and the vertical velocity field at 500 hPa diagnosed using Eqs. 5,6 from the Pangu-Weather forecast. Comparing the top and middle panels of Figure 8, it is apparent that while the diagnosed $w$ is unrealistic in regions with significant topography (i.e., where isobaric surfaces end up below ground), it Is a qualitatively good proxy for the forecasted $w$ in low lying areas and over the oceans. The other general feature that is apparent from inspection of Figure 8 is that the Pangu-Weather diagnosed w field (bottom panel) appears weaker, more diffuse than the ERA5 hindcast fields (both forecasted and diagnosed). This visual impression is quantitatively confirmed in Figure 9, where the evolution of the absolute value of the diagnosed vertical velocities averaged over the ocean are presented for the IFS forecasts, the ERA5 hindcasts and Pangu-Weather forecasts. Pangu-Weather vertical velocities are about 40% smaller than those of the ERA5 hindcasts (which have the same nominal spatial resolution) and about half in magnitude of those diagnosed from the IFS forecasts, which have higher resolution. Another notable aspect in this plot is the clear reduction of the intensity of vertical velocities in the Pangu-Weather forecasts during the first 24 hours, which is consistent with the reduction of the absolute value of the divergence-vorticity ratio shown in Figure 7.

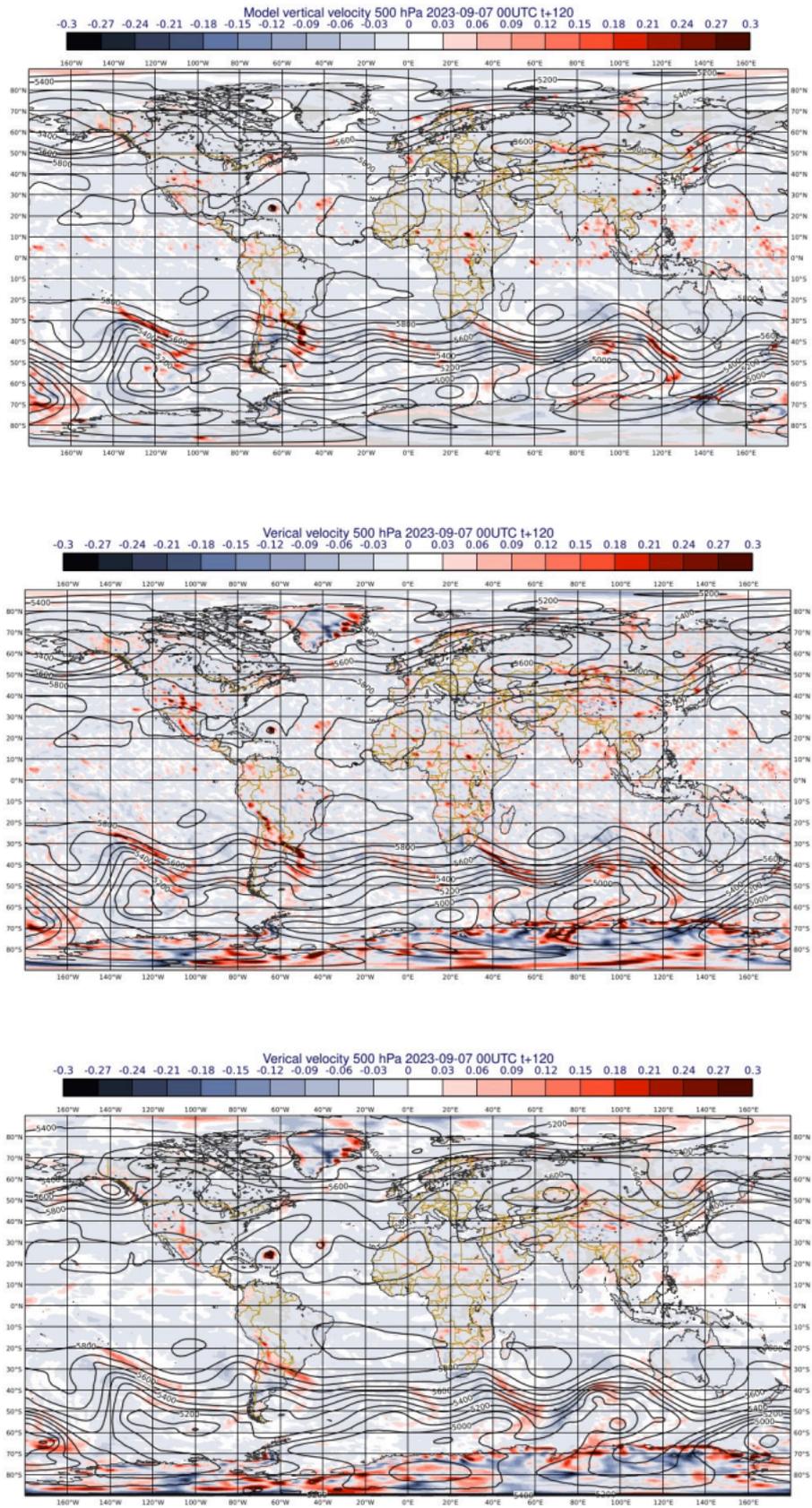

Figure 8: *Global plots of vertical velocities (shaded, units: m/s) at 500 hPa from: ERA5 hindcast (top panel); diagnosed from ERA5 hindcast horizontal wind components using Equations 5,6*

*(middle panel); diagnosed from Pangu-Weather forecast horizontal wind components using Equations 5,6 (bottom panel). Black isolines show geopotential height forecasts at 500 hPa from the corresponding systems.*

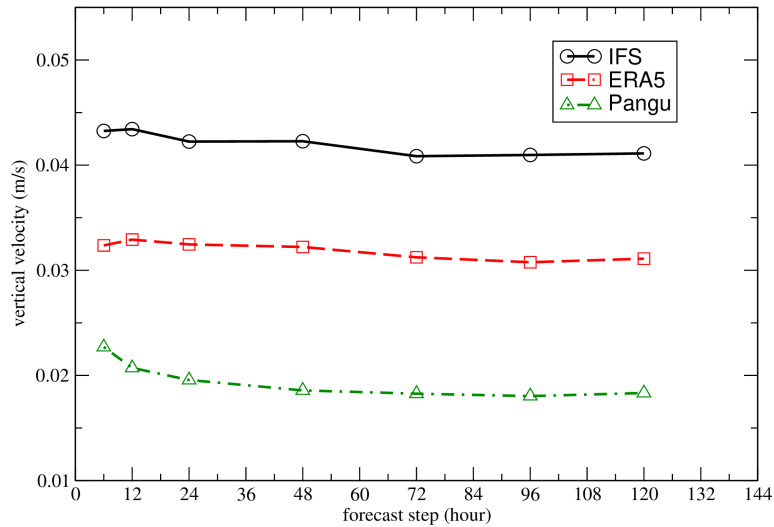

Figure 9: *Forecast evolution of the absolute value of the vertical velocity field diagnosed with Equations 5,6 for the ECMWF IFS model (continuous line); the ERA5 hindcasts (dashed line); and the Pangu-Weather forecast (dash-dot line). Values averaged over the sea and the period 2023-09-07 to 2023-9-10.*

All the three forecast maps presented in Figure 8 reveal the presence in the forecast of a developed tropical cyclone north of the Caribbean islands. This tropical cyclone corresponds to hurricane Lee, the strongest major hurricane of the 2023 Atlantic hurricane season at the time of writing (October 2023), and a category 3 hurricane with a 948 hPa minimum mslp at the verification time of the maps in Figure 8 (12 September 2023, 00 UTC). It is thus of interest to look into more detail of how Pangu-Weather handles this extreme weather phenomenon. A partial answer is provided in Figure 10, where we show a magnified view around the forecasted position of Hurricane Lee for the IFS, ERA5 and Pangu-Weather models. Note that the vertical velocities shown in all the plots in Figure 10 are obtained using Equations 5,6 on forecasted horizontal wind fields on standard pressure levels. While the relative shallowness of the Pangu-Weather forecasted tropical cyclone is coherent with the diagnostics presented in Figures 7-9, the general noisiness and lack of realism of the TC simulated in the Pangu-Weather forecast raise further concerns about the ability of this MLWP model to provide a physically consistent picture of the evolution of the atmosphere.

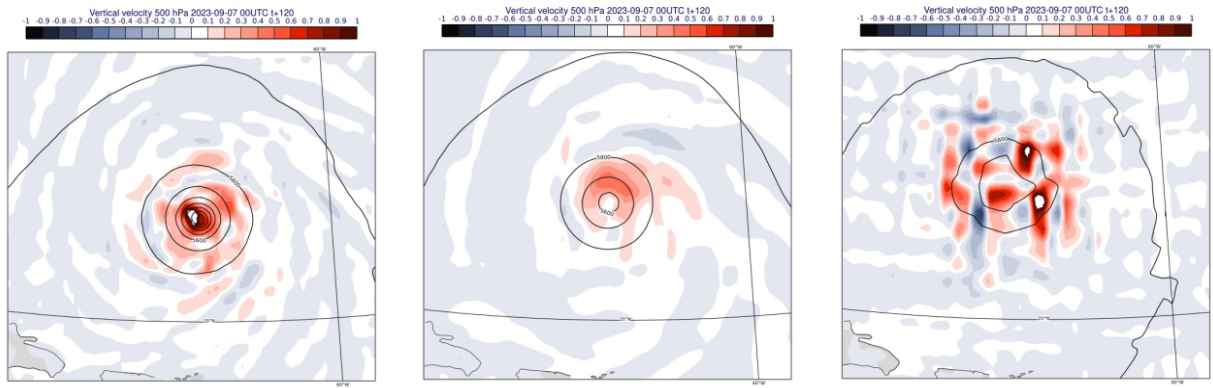

Figure 10: *Plots of t+120 hours forecast vertical velocities (shaded, units: m/s) at 500 hPa from: IFS (left panel); ERA5 hindcast (middle panel); and Pangu-Weather (right panel), diagnosed using Equations 5,6. Black isolines show geopotential height forecasts at 500 hPa from the corresponding systems. Forecasts valid on 2023-09-12, 00UTC.*

## 5. Forecast performance

The main purpose of this section is not to provide an exhaustive evaluation of the forecast skill of the Pangu-Weather model relative to the ECMWF IFS or EM operational products (see for example Ben-Bouallegue et al., 2023, for more detailed comparison) but to highlight how the characteristics of the models discussed in the previous sections may affect the results of standard deterministic verification. To start with, it is important to understand that a fair comparison involves using forecasts that have seen the same observational information in their initialisation. In the Pangu-Weather literature (Bi et al., 2022, 2023) the Pangu-Weather forecasts are initialised from ERA5 reanalysis fields and compared with ECMWF IFS operational forecasts. This is understandable as Pangu-Weather has been trained on ERA5 reanalysis fields and thus should perform best when ERA5 reanalyses are used for initialisation. On the other hand, the assimilation window used in the ERA5 analyses is 12 hours long while the assimilation window used for the operational ECMWF IFS and EM forecasts is 6 hours long (i.e., for the 12 UTC analyses, the ERA5 assimilation window is 09-21 UTC, the ECMWF IFS window is 09-15 UTC). This results in the ERA5 analyses using observations up to 6 more hours into the forecast range than ECMWF IFS forecasts, which translates into an approx. 6-hour advantage in terms of forecast skill for the model using ERA% initial conditions. This can be significant for many variables (e.g., Figure 11). To avoid this confounding factor, we present in the following results obtained by initialising Pangu-Weather from operational ECMWF IFS analyses, though we acknowledge that this could be slightly disadvantageous to Pangu-Weather performance.

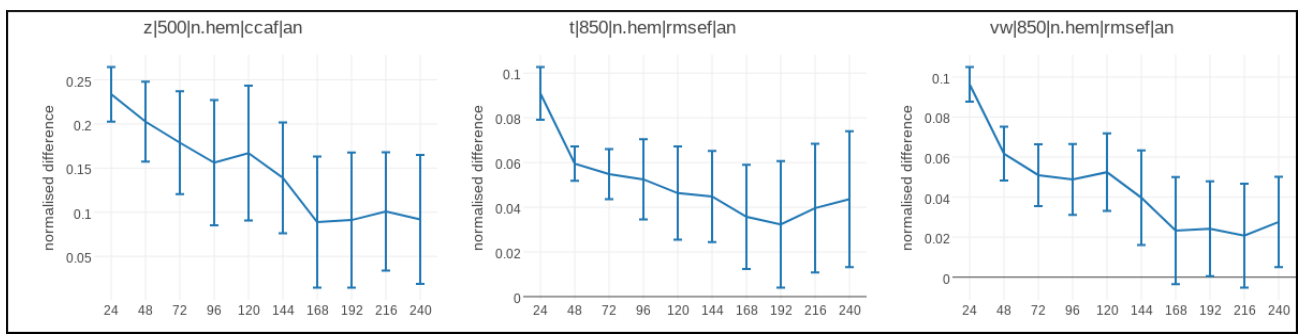

Figure 11: *Normalised forecast skill scores for ECMWF IFS initialised from a 6 hour ECMWF 4D-Var assimilation window and ECMWF IFS initialised from a 12 hour ECMWF 4D-Var assimilation window: Forecast anomaly correlation for geopotential at 500 hPa (left panel); forecast RMSE for temperature at 850 hPa (middle panel); forecast RMSE for wind speed at 850 hPa (right panel. Values above the zero line indicate superior skill of the ECMWF IFS run from 12 hours assimilation window analyses.*

In Figures 12 and 13 we present summary plots of forecast performance for a selection of variables (geopotential, temperature, wind). These results are consistent with the diagnostics on forecast activity presented in Sec. 3. There is broadly similar skill in RMSE terms in the short range (day 1-3) for all the models, consistent with the fact that all the models in the comparison show similar level of activity up to synoptic scale wavenumbers. In the medium-range (day 3-10) Pangu-Weather shows marginally improved performance over the ECMWF IFS, and significantly worse performance than the ECMWF ensemble forecast mean (EM). This behaviour is more apparent in the extra-tropics, where error growth is driven by loss of predictability of weather systems on synoptic scales, less in the tropics where errors have slower growth rates and are more influenced by large scale systematic errors.

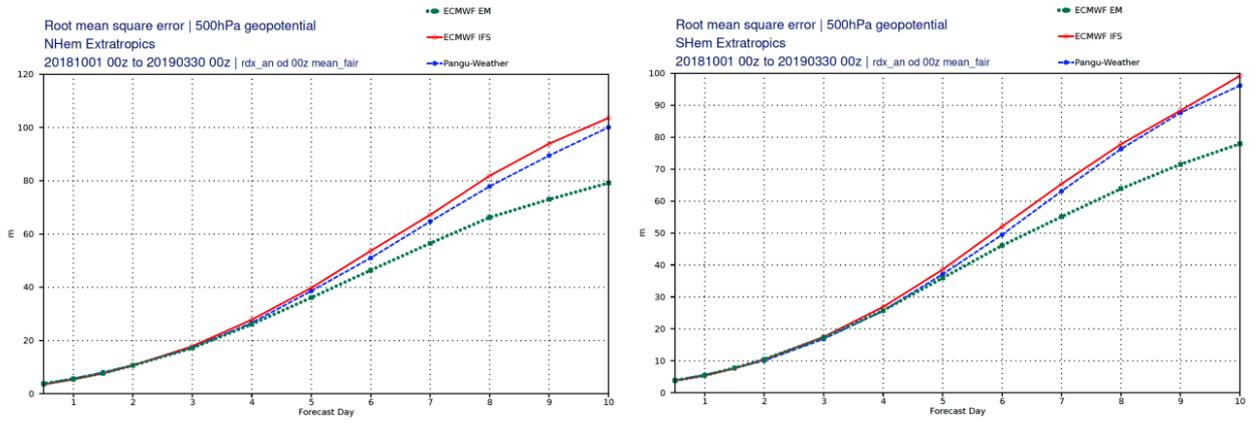

Figure 12: *RMS forecast error of 500hPa geopotential for the ECMWF IFS model (continuous line), Pangu-Weather (dash line) and ECMWF EM (dotted line) over the period 2018-10-01 to 2019-3-30, for the northern extra tropics (lat > 20 deg; left panel) and the southern extra tropics (lat < -20 deg; right panel). Verification against ECMWF analyses.*

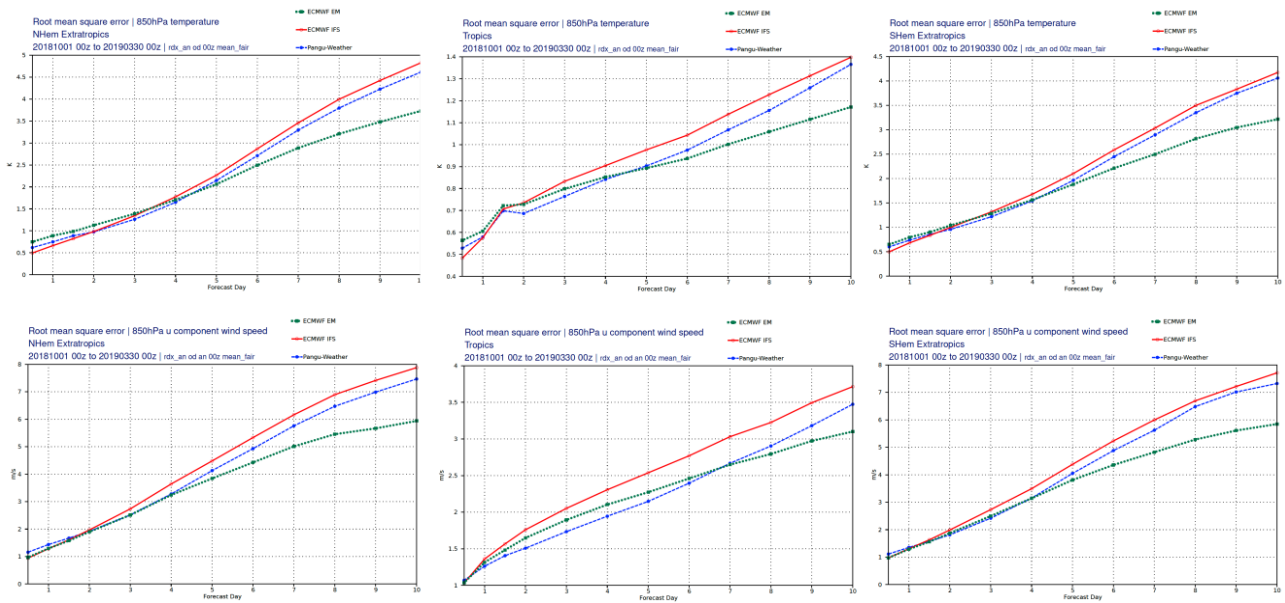

Figure 13: *RMS forecast error of 850 hPa temperature (top) and u-wind (bottom) for the ECMWF IFS model (continuous line), Pangu-Weather (dash line) and ECMWF EM (dotted line) over the period 2018-10-01 to 2019-3-30, for the northern extra tropics (lat > 20 deg; left panels), the tropics ( 20 deg >lat > -20; middle panels) and the southern extra tropics (lat < -20 deg; right panels). Verification against ECMWF analyses.*

**Discussion and Conclusions**

The field of ML weather forecasting has made huge progress in a very short (by traditional NWP development timelines) period of time, helped by the availability of high resolution, high accuracy training datasets like ERA5 and the more general progress in ML research and hardware and software tools. The best performing MLWP models routinely claim better performance than state-of-the-art traditional NWP models on a variety of deterministic forecast scores, while using orders of magnitude less computational power and energy than standard NWP models during their deployment phase (though they are completely reliant on reanalysis datasets which have themselves been assembled using orders of magnitude more computational resources and energy than any current operational NWP forecast). While these advantages appear compelling from an end-user perspective, it is important to realise some of the limitations inherent in the current generation of ML models, at least as much as they can be inferred from an analysis of the characteristics of an example ML model (Pangu-Weather). Pangu-Weather has been chosen because we believe it is broadly representative of the current generation of ML models and has been an early adopter of an open data and model policy. We plan to extend this analysis to other ML models, but we expect that main findings will generally hold.

One main finding of this analysis has been that Pangu-Weather (and by extrapolation, other current ML models) is not a general-purpose atmosphere simulator or, to use a terminology that has become popular in recent years, a ML-driven atmospheric digital twin. This is already apparent from the power spectra of Pangu-Weather forecasts when compared to those of the ERA5 analyses used for its generation and those of the ECMWF IFS model. Pangu-Weather forecast spectra show decreasing energy with increasing wavenumber (higher spatially resolved scales) and with increasing forecast lead time. This is in line with the common empirical observation that ML weather models produce progressively smoother, blurrier forecasts. What is possibly not so widely appreciated is that the shape and evolution of its forecast spectra imply that Pangu-Weather forecasts have limitations in representing fundamental dynamical balance relationships in the atmospheric motions, e.g., in the examples discussed here, those implied by geostrophic and ageostrophic flows and the ratio between divergent and rotational wind components. Additionally, the fact that these balances are not satisfied implies that other quantities that can be diagnosed or inferred from the mass field and the horizontal flow, e.g., vertical motion fields and, by extension, areas of precipitation/active weather, will also be unrealistic, which can be a limiting factor from a weather forecaster perspective. In a sense, this behaviour is not surprising, as ageostrophic and divergent flows are intrinsically less predictable than balanced and rotational flows, while the current generation of MLWP models are built to discover and reproduce predictable regressions between current and predicted states of the atmospheric training dataset.

Pangu-Weather, like all ML models considered here, is trained to minimise a L1/L2 loss function of forecast errors. This leads to the expectation that these models should better be considered estimators of the central moment of the forecast error pdf (median/mean). However, this interpretation is not completely straightforward. A comparison of the spectra of Pangu-Weather forecasts with those of the ECMWF operational ensemble forecast mean shows some discrepancies. In particular, Pangu-Weather forecasts do not present the signature drop in energy of the ECMWF EM at synoptic scales in the medium range (3-5 days), which is associated with the loss of predictability at these lead times/spatial scales due to the chaotic growth of initial and forecast uncertainties (Žagar, 2017), while they consistently show reduced forecast variability at smaller (sub-synoptic, mesoscale) spatial scales. This reduced forecast activity is also dependent on forecast lead time, more visibly during the first 24 hours, which implies significant heteroscedasticity in the distribution of forecast errors with forecast lead time. These results agree with recent work (Selz and Craig, 2023) documenting the inability of Pangu-Weather to produce realistic error growth from small-amplitude initial condition perturbations (i.e., lack of a "butterfly effect"). For these reasons, using effectively Pangu-Weather, and other similar ML models, in an ensemble forecast configuration may turn out to be more challenging than commonly anticipated, at least in terms of following the currently accepted paradigm of building forecast ensembles as collections of equiprobable realisations of physically consistent model trajectories.

The above considerations are also important in an objective evaluation of the forecast performance of Pangu-Weather and similar ML models. Forecast models with reduced variability and which do not present the standard upscale error growth of physics-based models (Selz and Craig, 2023) tend to perform better on deterministic forecast skill measures, especially at longer lead times ("double penalty" effect), and this is confirmed by results presented here. Whether this is a major reason of the good forecast skill of MLWP models, or only a contributing factor (for example, it is reasonable to expect these models to do a good job at correcting state-dependent systematic errors of the IFS) is a question that remains to be addressed. This also indicates that the intrinsic value of ML models will be better be evaluated on a more comprehensive set of measures in an ensemble forecasting context, where both the sharpness and reliability of the forecast pdf can be adequately assessed.

While they cannot be classified as atmospheric emulators/digital twins, ML models like Pangu-Weather can be better understood as forecast applications targeted at optimising specific aspects of forecast performance, i.e., minimising medium-range mean squared/absolute errors, over a range of representative atmospheric and near surface weather parameters. This is effectively a similar objective as that pursued by modern multivariate NWP post-processing techniques (Lakatos et al., 2023), with the obvious advantage that ML models regressions use the analysis state (or an ensemble of analyses) state as a predictor, instead of having to wait for the output of a deterministic or ensemble NWP forecast system. This can be both effective and efficient for various medium and extended-range user applications (Lam et al., 2022) where the main drivers of predictability are on synoptic or larger scales, standard NWP models are affected by significant systematic errors and producing physically consistent forecast states is not crucial for the end user. In fact, the ability of ML models to provide skilful forecasts on synoptic scales in the medium range suggests that similar tools could be effectively applied to different forecast ranges with a judicious choice of loss functions and training curricula. Current ML models are optimised to minimise errors in the 24-72 hour range. It is conceivable that for seasonal to climate prediction applications a loss function targeted at minimising errors on longer forecast ranges could prove effective in constraining error growth and reducing model drift on these longer time scales. At the opposite end of the forecast range spectrum, ML models trained to minimise errors on short (1-6 hour) timescales and with appropriate multivariate constraints in the loss function on forecast spectra could provide useful building blocks of an efficient MLWP ensemble prediction system able to realistically simulate the chaotic growth of initial and forecast uncertainties.

More generally, the discussion above and the results presented here highlight one of the main challenges for the next generation of data-driven ML prediction models, namely, how to produce forecasts that are skilful and at the same time dynamically and physically consistent at all relevant spatial scales. This will increase the interpretability and trustworthiness of the ML prediction models and ultimately encourage user acceptance of their outputs.

While it is clear that current ML models have a significant role to play in forecast applications, it is also important to keep in mind that they still inherently depend on a physics-based model and data assimilation system for both their training and their initialisation. The further development of the forecasting capabilities of the ML models is still fundamentally dependent on the undoubtedly slower paced but still necessary, methodical development of physics-based model and data assimilation systems.

**ACKNOWLEDGMENTS**

The Author would like to thank Matt Chantry (ECMWF) for making the Pangu-Weather forecast dataset available and for interesting and challenging discussions on the subject of Machine Learning applications to Weather and Climate Prediction. Gregory Hakim (Univ. of Washington), Linus Magnusson, Tony McNally, Andy Brown and Florian Pappenberger (ECMWF) provided insightful comments for an earlier version of this manuscript, which are gratefully acknowledged. The Author would also like to thank many colleagues at ECMWF and participants to the ECMWF-ESA Machine Learning Workshop series for enriching discussions on this fascinating topic.